\documentclass[myios]{article}

\usepackage{amssymb,amsbsy,amsmath,amsfonts,amssymb,amscd}
\usepackage{latexsym,euscript,exscale,epic,eepic,epsfig}
\usepackage[T1]{fontenc}
\usepackage{times}
\usepackage[all]{xy}
\usepackage{graphicx} 

\usepackage{epsfig}
\usepackage[english]{babel}
\usepackage{amsmath}
\usepackage{amsfonts}
\usepackage{amssymb}

\usepackage{a4wide}
\usepackage{graphicx}
\usepackage{epsfig}
\usepackage{amsmath}
\usepackage{amsfonts}
\usepackage{amssymb}
\usepackage{latexsym}

\newtheorem{theorem}{Theorem}[section]
\newtheorem{theorem*}{Theorem}

\newtheorem{definition}{Definition}[section]
\newtheorem{definition*}{Definition}

\newtheorem{proposition}{Proposition}[section]

\newtheorem{remark}{Remark}[section]

\newenvironment{proof}%
{
\noindent \textbf{Proof.} \small}%

\def\R{\mathbb{R}}

\def\N{\mathbb{N}}

\def\mB{\mathcal{B}}

\def\fB{\mathfrak{B}}

\def\1{\mbox{1\hspace{-.15em}\small{l}}}
 \def\i1{\int_{0}^{1}}

\def\rkhs{r.k.h.s}

             \def\Un{{\mathrm{{1\hskip-2.6pt I}}}}
 \def\R{{\mathrm{I\hskip-2.2pt R}}}
\def\dbN{{\mathrm{I\hskip-2.2pt N}}}

\def\w{{\mathbf{\weight}}}

\def\x{{\mathbf{x}}}
\def\y{{\mathbf{y}}}

\def\dbR{{\mathrm{I\hskip-2.2pt R}}}
\def\clL{{\cal{L}}}
\def\clX{{\cal{X}}}
\def\clY{{\cal{Y}}}
\def\clF{{\cal{H}}}
\def\clH{{\cal{H}}}
\def\clM{{\cal{M}}}

\def\w{{\mathbf{w}}}
\def\x{{\mathbf{x}}}
\def\y{{\mathbf{y}}}

\def\dbR{{\mathrm{I\hskip-2.2pt R}}}
\def\dbN{{\mathrm{I\hskip-2.2pt N}}}

\def\vect{{\mbox{Vect}}}

\begin{document} 
\title{Functional learning through kernel}
\author{
  St{\'e}phane Canu,  Xavier Mary and Alain Rakotomamonjy \\ PSI \\ INSA  de Rouen,\\
  St Etienne du Rouvray, France \\
  {\tt stephane.canu,xavier.mary,alain.rakotomamonjy@insa-rouen.fr }
}
\maketitle

\begin{abstract}
This paper reviews the functional aspects of  statistical learning theory.
The  main point under consideration is the nature of the hypothesis set 
when no prior information is available but  data.
Within this framework 
we first discuss about the hypothesis set:
it is a vectorial space, it is a set of pointwise defined functions, 
and the evaluation functional on this set is a continuous mapping.
Based on these principles an original theory is developed
generalizing the notion of reproduction kernel Hilbert space to non hilbertian sets.  
Then it is shown that the hypothesis set of any learning machine
has to be a generalized reproducing set.
Therefore, thanks to a general ``representer theorem'',
the solution of the learning problem
is still a linear combination of a kernel.
Furthermore, a way to design these kernels is given.
To illustrate this framework some examples of such reproducing sets and kernels are given.

\end{abstract}
\sloppy
\section{Some questions regarding machine learning}

Kernels and in particular Mercer or reproducing kernels play a
crucial role in statistical learning theory and functional estimation. 
But very little is known about the associated hypothesis set,
the underlying functional space where learning machines  look for the solution.
How to choose it? How to build it? What is its relationship with regularization?
%
%
%
The machine learning community has been interested in tackling the problem the other way round.
For a given learning task, therefore for a given hypothesis set,
is there a learning machine capable of learning it?
The answer to such a question allows to distinguish between learnable and non-learnable problem.
The remaining question is: is there a learning machine
capable of learning any learnable set.

\noindent
We know since \cite{poggio89theory} that learning is closely related to the approximation theory,
to the generalized spline theory, to regularization 
and, beyond, to the notion of reproducing kernel Hilbert space ($\rkhs$).
This framework is based on the minimization of the empirical cost plus a stabilizer ({\em i.e.} a norm is some Hilbert space).
Then, under these conditions, 
the solution to the learning task is a linear combination of some positive kernel 
whose shape depends on the nature of the stabilizer.
This solution is characterized by strong and nice properties such as universal consistency.

\noindent
But within this framework there remains a gap between 
theory and  practical solutions implemented by practitioners.
For instance, in $\rkhs$, kernels are positive. Some practitioners use hyperbolic tangent kernel 
 $\mbox{tanh}(\w^\top \x+w_0)$ while it  is not a positive kernel: but it works.
Another example is given by practitioners using non-hilbertian framework.
The sparsity upholder uses absolute values 
such as  $\int |f| d\mu$ or $\sum_j |\alpha_j|$: these are $L^1$ norms. They are not hilbertian.
Others escape the hilbertian approximation orthodoxy by introducing prior knowledge ({\em i.e.} a stabilizer)
through information type criteria that are not norms.

\noindent
This paper aims at revealing  some underlying hypothesis of the learning task
extending the  reproducing kernel Hilbert space framework.
To do so we begin with reviewing some 
learning principle. 
We will stress that the hilbertian nature of the hypothesis  set is not necessary
while the reproducing  property is.
This leads us to define a non hilbertian framework for  reproducing  kernel
allowing non positive kernel, non-hilbertian norms and other kinds of stabilizers.

\noindent
The paper is organized as follows.
The first point is to establish the three basic principles of learning.
Based on these principles and
 before  entering the non-hilbertian framework,
it appears necessary to recall some basic elements of the theory of reproducing kernel Hilbert space 
and how to build them from non reproducing Hilbert space.
Then the construction of non-hilbertian reproducing space is presented
by replacing the dot (or inner) product by a more general duality map.
This implies  distinguishing between two different sets put in duality, 
one for hypothesis and the other one for measuring. 
In the hilbertian framework these two sets are merged in a single Hilbert space.

\noindent
But before  going into technical details 
we think it advisable to review the use of $\rkhs$ in the learning machine community.

\section{$\rkhs$ perspective}

\subsection{Positive kernels}
The interest of $\rkhs$ arises from its associated kernel.
As it were, a $\rkhs$ is a set of functions entirely defined by a kernel function.
A Kernel may be characterized as a function from $\clX \times \clX $ to $\dbR$   
(usually $\clX \subseteq R^d$). 
Mercer \cite{Mercer09} first establishes some remarkable properties of a particular class of kernels: 
positive kernels defining an integral operator. 
These kernels have to belong to some functional space 
(typically $L^2(\clX \times \clX)$, the set of square integrable functions on $\clX \times \clX$)
so that the associated integral operator is compact.
The positivity of  kernel $K$  is defined as follows:
$$
K(x,y) \mbox{ positive} \;\; \Leftrightarrow \;\; \forall f \in L^2, \;\; \langle \langle K,f\rangle_{L^2},f\rangle_{L^2} \geq 0
$$
where $\langle.,.\rangle_{L^2} $ denotes the dot product in $L^2$.
Then, because it is compact, the kernel operator admits a countable spectrum 
	and thus the kernel can be decomposed.
Based on that, the work by Aronszajn \cite{Aronszajn50} can be presented as follows.
Instead of defining the kernel operator from $L^2$ to $L^2$ Aronszajn focuses on the $\rkhs$ $H$ embeded with its dot product $\langle .,.\rangle_H$.
In this framework the kernel  has to be a pointwise defined function.
The positivity of  kernel $K$  is then defined as follows:
\begin{equation} \label{eq:posKerAron}
K(x,y) \mbox{ positive} \;\; \Leftrightarrow \;\; \forall g \in H, \;\;  \langle \langle K,g\rangle_H,g\rangle_{H}  \geq 0
\end{equation} 
Aronszajn first establishes a bijection between kernel and $\rkhs$.
Then L. Schwartz \cite{Schwartz64} shows that this was a particular case of a more general situation.
The kernel doesn't have to be a genuine function.
He generalizes the notion of positive kernels to weakly continuous linear application from 
the dual set $E^*$ of a vector space $E$ to itself.
To share interesting properties the kernel has to be positive in the following sense:
$$
K \mbox{ positive} \;\; \Leftrightarrow \;\;  \forall h \in E^* \;\;  ((K(h),h)_{E,E^*} \geq 0
$$
where $(.,.)_{E,E^*} $ denotes the duality product between $E$ and its dual set $E^*$.
The positivity is no longer defined in terms of scalar product.
But there is still a bijection between positive Schwartz kernels and Hilbert spaces.

\smallskip
\noindent
Of course this is only a short part of the story.
For a detailed review on $\rkhs$ and a complete literature survey see  \cite{Atteia92,Saitoh88}.
Moreover some authors consider non-positive kernels.
A generalization to Banach sets has been introduced \cite{Att81} 
within the framework of the approximation theory. 
Non-positive kernels have been also introduced in Kre\u{i}n spaces 
as the difference between two positive ones (\cite{Alpay87} and  \cite{Schwartz64} section 12).

\subsection{$\rkhs$  and learning in the literature}

The first contribution of $\rkhs$ to the statistical learning theory  
is the regression spline algorithm.
For an overview of this method see Wahba's book \cite{wahba_spline}.
In this book %
two important hypothesis regarding the application of the $\rkhs$ theory to statistics are stressed.
These are
the nature of pointwise defined functions
and the continuity of the evaluation functional\footnote{These definition 
are formaly given section 3.5, definition 3.1 and equation (\ref{eq:cont})}.
An important and general result in this framework is the so-called representer theorem \cite{kimeldorf_wahba}.
This theorem states that the solution of some class of approximation problems
 is a linear combination of a kernel evaluated at the training points. 
But only applications in one or two dimensions are given.
This is due to the fact that, in that work, the way to build $\rkhs$ 
was based on some derivative properties.
For practical reason only low dimension regressors were considered by this means.

\noindent
Poggio and  Girosi extended the framework to large input dimension
by introducing radial functions through regularization operator \cite{poggio89theory}.
They show  how to build such a kernel as the green functions
 of a differential operator defined by its Fourier transform.

\noindent
Support vector machines (SVM) perform another important link between 
kernel, sparsity and bounds on the generalization error \cite{vapnik_nSLT}.
This algorithm is based on  Mercer's theorem and on the relationship between kernel and dot product. 
It is based on the ability for positive kernel 
to be separated and decomposed according to some generating functions.
But to use Mercer's theorem the kernel has to define a compact operator.
This is the case for instance when it belongs to $L^2$ functions defined on a compact domain.

\noindent
Links between green functions, SVM and reproducing kernel Hilbert space were introduced in
 \cite{girosi_sparse} and \cite{smola98from}.
The link between $\rkhs$ and bounds on a compact learning domain 
has been presented in a mathematical way  by Cucker and Smale \cite{CuckerSmale01}.

\medskip
\noindent
Another important application of $\rkhs$ to learning machines 
comes from the bayesian learning community.
This is due to the fact that, in a probabilistic framework, 
a positive kernel is seen as a covariance function associated to a gaussian process. 

\section{Three principles on the nature of the hypothesis set}

\subsection{The learning problem}

 A supervised learning problem 
is defined by a learning domain ${\cal X} \subseteq \dbR^d$
 where $d$ denotes the number of explicative variables,
  the learning codomain  ${\cal Y} \subseteq \dbR$ 
and a $n$ dimensional  sample $\{(\x_i,y_i), i=1,n\}$: the training set.

Main stream formulation of the learning problem 
considers the loading of a learning machine based on empirical data
 as the minimization of a given criterion with respect to some hypothesis
lying in a  hypothesis set $\clF$. 
In this framework hypotheses are functions $f$ from $\clX$ to $\clY$
and the hypothesis space $\clF$ is a functional space.
$$
\mbox{ Hypothesis }  H_1: \clF \mbox{ is a functional vector space}
$$
Technically a convergence criterion is needed in $\clF$, {\em i.e.} $\clF$ has to be embedded with a topology.
In the remaining, we will always assumed $\clF$ to be a convex topological vector space.

\medskip
\noindent
Learning is also the minimization of some criterion. 
Very often the criterion to be minimized contains two terms.
The first one, $C$, represents the fidelity of the hypothesis with respect to  data
while $\Omega$, the second one, represents the compression 
required to make a difference between memorizing and learning.
Thus the learning machine solves the following minimization problem:
\begin{equation}\label{eq:cout}
\min_{f \in \clF} \; C(f(x_1),...,f(x_n),\y) + \Omega(f) 
\end{equation}
The fact is, while writing this cost function, we implicitly assume that the 
value of function $f$ at any point $x_i$ is known.
We will now discuss the important consequences this assumption has 
on the nature of the hypothesis space $\clF$.

\subsection{The evaluation functional}

By writing $f(x_i)$ we are assuming that function $f$ can be evaluated at this point.
Furthermore if we want to be able to use our learning machine to make a prediction for a given input $x$, 
 $ f(x)$ has to exist for all $x \in \clX$: we want pointwise defined functions.
This property is far from being shared by all functions.
For instance  function $sin(1/t)$ is not defined in 0.
 Hilbert space $L^2$ of square integrable functions 
is a quotient space of functions defined only almost everywhere ({\em i.e.} 
not on the singletons $\{x\}, x \in \clX$).
%
%
 $L^2$ functions are not pointwise defined because  the  $L^2$ elements are equivalence classes.

To formalize our point of view we need to define
$\dbR^\clX$ as the set of all pointwise defined functions from $\clX$ to $\dbR$.
For instance when $\clX=\dbR$ all finite polynomials (including constant function) belong to $\dbR^\clX$.
%
We can lay down our second principle:
$$
\mbox{ Hypothesis }  H_2: \clF \mbox{ is a set of pointwise defined function ({\em i.e.}  a subset of $\dbR^\clX$)}
$$

Of course this is not enough to define  a hypothesis set properly
and at least another fundamental property is required.

\subsection{Continuity of the evaluation functional}

The pointwise evaluation of the hypothesis function is not enough.
We want also the pointwise convergence of the hypothesis.
If two functions are closed in some sense 
we don't want them  to disagree  on any point. 
Assume $t$ is our unknown target function to be learned. 
For a given sample of size $n$ a learning algorithm provides a hypothesis $f_n$.
Assume this hypothesis converges in some sense to the target hypothesis.
Actually the reason for hypothesis $f_n$ is that it will be used to predict the value of $t$ 
at a given $x$. For any $x$ we want $f_n(x)$ to converge to $t(x)$ as follows:
$$
f_n \overset{\clF}{\longrightarrow} t
 \Longrightarrow
\forall x \in \clX , \; f_n(x) \overset{\dbR}{\longrightarrow} t(x)
$$
We are not interested in global convergence properties but in local convergence properties.
Note that it may be rather dangerous to define a learning machine without this property.
%
Usually the topology on $\clF$ is defined by a norm.
Then the pointwise convergence can be restated as follow:
\begin{equation} \label{eq:cont}
\forall x \in \clX ,\; \exists M_x \in \dbR^+ \mbox{ such that } |f(x) - t(x)| \; \leq \; M_x \; ||f-t||_\clH
\end{equation}
At any point $x$, the error can be controlled.

It is interesting to restate this hypothesis with the evaluation functional
\begin{definition}  the evaluation functional 
$$
\begin{array}{llll}
\delta _x : & \clF &\longrightarrow & \dbR\\
            & f    &\longmapsto     & \delta_x f = f(x) \\
\end{array}
$$
\end{definition} 
Applied to the evaluation functional our prerequisite  of pointwise convergence is equivalent to its continuity.
$$
\mbox{ Hypothesis }  H_3:\mbox{ the evaluation functional is continuous on } \clF 
$$ 
Since the evaluation functional
 is linear and continuous, it belongs to the topological dual of $\clF$.
We will see that this is the key point to get the reproducing property.

\smallskip
\noindent
Note that the continuity of the evaluation functional does not necessarily imply
 uniform convergence. But in many practical cases it does.
To do so one additional  hypothesis is needed, the constants $M_x$ have to be bounded: 
$ \sup_{x \in \clX} M_x < \infty $.
For instance this is the case when the learning domain $\clX$ is bounded.
Differences between uniform convergence and evaluation functional continuity
is a deep and important topic for learning machine but out of the scope of this paper.

\subsection{Important consequence}
To build a learning machine we do need to choose our hypothesis set as a reproducing space
to get the pointwise evaluation property and the continuity of this evaluation functional.
But the Hilbertian structure  is not necessary.
Embedding  a set of functions with the property of continuity of the evaluation functional
has  many interesting consequences. 
The most useful one in the field of learning machine
 is the existence of a kernel $K$,
a two-variable function with  generation property\footnote{this property 
means that the set of all finite linear combinations of the kernel is dense in $\clF$. 
See proposition 4.1 for a more precise statement.}:
$$
\forall f \in \clF, \; \exists \ell \in \dbN, \; (\alpha_i)_{i=1,\ell} \mbox{ such that } \; 
f(x) \approx \sum_{i=1}^\ell \alpha_i K(x,x_i)
$$
$I$ being a finite set of indices. 
Note that for  practical reasons $f$ may have a different representation.

If the evaluation set is also a Hilbert space (a vector space embedded with a dot product)
it is a reproducing kernel Hilbert space ($\rkhs$).
%
Although not necessary, $\rkhs$ are widly  used for learning
because they have  a lot of nice practical properties.
Before  moving on more general reproducing sets,
let's review the most important properties  of $\rkhs$ for learning. 

\subsection{ $\dbR^\clX$ the set of the pointwise defined functions on $\clX$}

\noindent In the following, the function space of the pointwise defined functions
 $\R^{\clX}=\{f:\clX \to \R\}$ will be seen as a topological vector space embedded with the
topology of simple convergence.

\noindent
$\dbR^\clX$ will be put in duality with $\dbR^{[\clX]}$ 
the set of all functions on $\clX$ equal to zero everywhere except  on a 
finite subset $\{x_i , i\in I\}$ of $\clX$.
Thus all functions belonging to $\dbR^{[\clX]}$ can be written in the following way:
$$
g \in \dbR^{[\clX]} \Longleftrightarrow \exists \left\{\alpha_i\right\},{i=1,n} \mbox{ such that } 
g(x) = \sum_i \alpha_i \Un_{x_i}(x)
$$
were the indicator function $\Un_{x_i}(x)$  is 
null everywhere except on $x_i$ where it is equal to one.
$$
\forall x\in \clX \quad \Un_{x_i}(x) = 0 \mbox{ if } x \neq x_i \mbox{ and } 
 \Un_{x_i}(x) = 1 \mbox{ if } x = x_i 
$$
Note that the indicator function is closely related to the evaluation functional since they are in bijection through:
$$\forall f \in \dbR^\clX, \forall x \in \clX, \quad \delta_x(f) = \displaystyle \sum_{y \in \clX} \Un_{x}(y)f(y) = f(x)$$ 
But formally, $\left(\dbR^\clX\right)'=\mbox{span}\{\delta_x\}$ is a set of linear forms while $\dbR^{[\clX]}$ is a set of pointwise defined functions.

\section{Reproducing Kernel Hilbert Space ($\rkhs$)}

\begin{definition}[Hilbert space]
A vector space $H$ embedded with the positive definite dot product $\langle .,. \rangle_H$  
is a Hilbert space if it is complete for the induced norm $\|f\|_H^2=\langle f,f\rangle_H$ 
 ({\em i.e.} all Cauchy sequences converge in $H$). 
\end{definition} 
For instance $\dbR^n$,  
 ${\cal{P}}_k$ the set of polynomials of order lower or equals to $k$, $L^2$, 
 ${\ell^2}$ the set of square sumable sequences seen as functions on $\dbN$ 
are Hilbert spaces.
$L^1$ and the set of bounded functions $L^\infty$ are not.

\begin{definition}[reproduction kernel Hilbert space ($\rkhs$)] 
A Hilbert space $(\clF, \langle .,. \rangle_\clF)$ is a $\rkhs$ 
    if it is defined  on $\dbR^\clX$  (pointwise defined functions)
and if the evaluation  functional is continuous on $H$ 
(see the definition of continuity equation \ref{eq:cont}).
\end{definition} 
For instance $\dbR^n$, ${\cal{P}}_k$  as any finite dimensional set of genuine functions are $\rkhs$. 
 ${\ell^2}$  is also a $\rkhs$. 
The Cameron-Martin space defined example 8.1.2 is a $\rkhs$ 
while  
$L^2$ is not because it is not a set of pointwise functions.

\begin{definition}[positive kernel] 
A function from $\clX \times \clX$ to $\dbR$ is a positive kernel 
if it is symmetric  and if for any 
 finite subset $\{x_i\}, i=1,n$ of $\clX$
and any sequence of scalar $\{\alpha_i\}, i=1,n$
$$
\sum_{i=1}^n \sum_{j=1}^n  \; \alpha_i \alpha_j K(x_i,y_j) \; \geq \; 0
$$
\end{definition} 
This definition is equivalent to Aronszajn definition of positive kernel given equation (\ref{eq:posKerAron}).

\begin{proposition} [bijection between $\rkhs$ and Kernel] Corollary  of proposition 23 in \cite{Schwartz64} 
and theorem 1.1.1 in \cite{wahba_spline}.
There is a bijection between the set of all possible $\rkhs$
and the set of all positive kernels.
\end{proposition}

\enlargethispage{-3cm}

\begin{proof}
\begin{itemize}
\item[$\Rightarrow$] from $\rkhs$  to Kernel. 
Let $(\clF, \langle . , . \rangle_\clF)$ be a $\rkhs$. By hypothesis the evaluation functional 
$\delta_x$ is a continuous linear form so that it belongs to the topological dual of $\clF$. 
Thanks to the Riesz theorem
we know that for each $x \in \clX$ there exists a function $K_x(.)$ belonging to $\clF$ such that
for any function $f(.) \in \clF$:
$$
\delta_x(f(.)) = \langle K_x(.),f(.) \rangle_\clF
$$
$K_x(.)$ is a function from $\clX \times \clX$ to $\dbR$ and thus can be written as 
a two variable function $K(x,y)$.
This function is symmetric  and positive since, for any real finite sequence $\{\alpha_i\},i=1,\ell$, 
$\sum_{i=1}^\ell \alpha_i K(x,x_i) \in \clF$, we have:
$$
\begin{array}{lll}
\|\sum_{i=1}^\ell  \alpha_i K(.,x_i)\|_\clF^2 &=& \langle \sum_{i=1}^\ell  \alpha_i K(.,x_i),\sum_{j=1}^\ell  \alpha_j K(.,x_j)\rangle_\clF \\
             &=& \displaystyle \sum_{i=1}^\ell  \sum_{j=1}^\ell  \alpha_i \alpha_j K(x_i,x_j)
\end{array}
$$

\item[$\Leftarrow$]from kernel to $\rkhs$.
For any couple $(f(.),g(.))$   of $\dbR^{[\clX]}$  
(there exist two finite sequences $\{\alpha_i\} i=1,\ell$ and $\{\beta_j\},j=1,m$ 
and two sequence of $\clX$ points $\{x_i\} i=1,\ell$, $\{y_j\},j=1,m$
 such that $f(x) = \sum_{i=1}^\ell  \alpha_{i=1}^\ell  \Un_{x_i}(x)$ 	and 
$g(x) = \sum_{j=1}^m \beta_j \Un_{y_j}(x)$) 
we define the following bilinear form:
$$
\langle f(.),g(.) \rangle_{[\clX]} =  \displaystyle \sum_{i=1}^\ell  \sum_{j=1}^m  \alpha_i \beta_j K(x_i,y_j)
$$
Let $\clH_0 = \{f \in \dbR^{[\clX]} ;|\; \langle f(.),f(.)\rangle_{[\clX]} = 0\} $.
$\langle .,.\rangle _{[\clX]}$ defines a dot product on the quotient set  $\dbR^{[\clX]}/\clH_0$.
Now let's define $\clH$ as the  $\dbR^{[\clX]}$ completion for the corresponding norm.
$\clH$ is a $\rkhs$ with kernel $K$ by construction. 
 
 \end{itemize}
\end{proof}

%
%
\begin{proposition}[from basis to Kernel]
Let $\clF$ be a  $\rkhs$. Its  kernel $K$ can be written:
$$
K(x,y) = \sum_{i\in I} \; e_i(x)\;e_i(y)
$$
for all orthonormal basis $\{e_i\}_{i\in I}$ of $\clF$, 
$I$ being a set of indices possibly infinite and non-countable.
\end{proposition}
\begin{proof}
$K\in \clF$ implies there exits a real sequence $\{\alpha_i\}_{i\in I}$ such that 
$K(x,.) = \sum_{i\in I} \alpha_i e_i(x)$. Then for all $e_i(x)$ element of the orthonormal basis:
$$
\begin{array}{lllll}
&\langle K(.,y) , e_i(.) \rangle_\clF  &=& e_i(y)  \qquad & \mbox{because of } 
K \mbox{ reproducing property} \\
\mbox{ and }& \langle K(.,y) , e_i(.) \rangle_\clF  &=& \langle \sum_{j\in I} 
\alpha_j e_j(.) , e_i(.) \rangle_\clF   &  \\
& &=&  \sum_{j\in I} \alpha_j \langle e_j(.) , e_i(.) \rangle_\clF   & \\
& &=&   \alpha_i &\mbox{ because } \{e_i\}_{i \in I} \mbox{ is an orthonormal basis }
\end{array}
$$
by identification we have $\alpha_i = e_i(y)$.
\end{proof}

\begin{remark}
Thanks to this results it is also possible to associate to any positive kernel a basis,
 possibly uncountable.
Consequenty to proposition 4.1 we now how to associate a $\rkhs$ to any positive kernel 
 and we get the result because every Hilbert space admit an orthonormal basis.
\end{remark}
\smallskip
\noindent
The fact that the basis is countable or uncountable 
(that the corresponding $\rkhs$ is separable or not) 
has no consequences on the nature of the hypothesis set (see example 8.1.7).
Thus Mercer kernels are a particlar case of a more general situation 
since every Mercer kernel is positive in the Aronszajn sense (definition 4.3)
while the converse is false.
Consequenty, when possible functionnal formulation is preferible to kernel formulation 
of learning algorithm.

\section{Kernel and kernel operator}

\subsection{How to build $\rkhs$?}

It is possible to build  $\rkhs$ from a $L^2(G,\mu)$ Hilbert space
where $G$ is a set (usualy $G=\clX$) and $\mu$ a measure.
To do so, an operator $S$ is defined to map $L^2$ functions onto 
the set of the pointwise valued functions $\dbR^\clX $.
A general way to define such an operator
consists in remarking that the scalar product  performs such a linear mapping.
Based on that remark this operator is built from a family $\Gamma_x$ 
of $L^2(G,\mu)$  functions when $x \in \clX$ in the following way:
\begin{definition}[Carleman operator]
Let $\Gamma = \{\Gamma_x, x \in \clX\}$ be a family of $L^2(G,\mu)$ functions. 
The associated Carleman operator $S$ is 
$$
\begin{array}{llll}
S : & L^2 &\longrightarrow & \displaystyle  \dbR^\clX \\
    & f   &\longmapsto     & g(.) = (Sf)(.)  = \langle \Gamma_{(.)} , f \rangle_{L^2}  
 = \displaystyle  \int_{G} \Gamma_{(.)} \; f \; d\mu 
\end{array}
$$
\end{definition}
That is to say $\forall x \in \clX, \; g(x) = \langle \Gamma_{x} , f \rangle_{L^2}$.  
To make apparent the bijective restriction of $S$ it is convenient to factorize it as follows:
\begin{equation}\label{eq:T}
S :  L^2 \longrightarrow L^2/\mbox{Ker}(S)  \overset{T}{\longrightarrow} \mbox{Im}(S) 
\overset{i}{\longrightarrow}  \displaystyle  \dbR^\clX 
\end{equation}
where $L^2/\mbox{Ker}(S)$ is the quotient set, $T$ the bijective restriction of $S$ and $i$  
the cannonical injection.

\smallskip
\noindent
This class of integral operators is known as Carleman operators \cite{Targon67}.
Note that this operator unlike Hilbert-Schmidt operators 
need not  be compact neither bounded.
But when $G$ is a compact set or when $\Gamma_x \in L^2(G \times G)$
(it is a square integrable function with respect to both of its variables)
$S$ is a  Hilbert-Schmidt operator. As an illustration of this property,
see the gaussian example on $G= \clX =\dbR$ in table \ref{table:gamma}. 
In that case $\Gamma_x(\tau) \not\in L^2(\clX \times \clX)$\footnote{
To clarify the not so obvious notion of pointwise defined function,
whenever possible, we use the notation $f$ when the function is not a pointwise defined function
and $f(.)$ denotes $\dbR^\clX$ functions. Here $\Gamma_x(\tau)$ is a pointwise defined function 
with respect to variable $x$ but not with respect to variable $\tau$. Thus, whenever possible, 
the confusing notation $(\tau)$ is omitted.}.
\begin{proposition}[bijection between Carleman operators and the set of $\rkhs$]
- Proposition 21 in \cite{Schwartz64} or theorems 1 and 4 in \cite{Saitoh88}. 
Let $S$ be a Carleman operator.
 Its image set $\clF = \mbox{Im}(S)$ is a \rkhs.
If $\clF$ is a  $\rkhs$ 
there exists a measure $\mu$ on some set $G$ 
 and a Carleman operator $S$ on $L^2(G,\mu)$ 
such that $\clF = \mbox{Im}(S)$. 
\end{proposition}

\begin{proof}
\begin{itemize}
\item[$\Rightarrow$]
Consider $T$ the bijective restriction of $S$ defined in equation (\ref{eq:T}).
 $\clF = \mbox{Im}(S)$  can be embedded  with the induced dot product defined as follows:
$$
\begin{array}{lllll}
\forall g_1(.),g_2(.) \in \clF^2,\quad  \langle g_1(.),g_2(.) \rangle_{\clF} 
&=& \langle T^{-1}g_1,T^{-1}g_2 \rangle_{L^2} \\
&=&  \langle f_1,f_2 \rangle_{L^2}  \qquad \qquad 
\mbox{ where } g_1(.) = T f_1 \mbox{ and } g_2(.) = T f_2
\end{array}
$$
With respect to the induced norm, $T$ is an isometry.
To prove $\clF$ is a $\rkhs$, we have to check the continuity of the evaluation functional.
This works as follows: 
$$
\begin{array}{lllll}
g(x) &=& \displaystyle \left(Tf\right)(x) &&\\
&=& \langle \Gamma_x, f \rangle_{L^2} &\leq& \| \Gamma_x\|_{L^2} \;\| f\|_{L^2}\\
& &                                        &\leq& \; M_x  \; \; \| g(.) \|_\clF
\end{array}
$$
with $M_x = \| \Gamma_x\|_{L^2} $. In this framework $\clF$ reproducing kernel $K$ verifies $S \Gamma_x = K(x,.)$.
It can be built based on $\Gamma $:
$$
\begin{array}{lll}
K(x,y) &=& \langle K(x,.) , K(y,.) \rangle_{\clF} \\
			&=& \langle \Gamma_x , \Gamma_y \rangle_{L^2} 
\end{array}
$$
\item[$\Leftarrow$]
Let $\{e_i\},i\in I$ be a $L^2(G,\mu)$ orthonormal basis and $\{h_j(.)\},j \in J$ an orthonormal basis of $\clF$.
We admit there exists a couple (G,$\mu$) such that $\mbox{card}(I) \geq \mbox{card}(J)$
(take for instance the counting measure on the suitable set).
Define $\Gamma_x = \sum_{j\in J} h_j(x) e_j$ as a $L^2$ family. 
Let $T$ be the associated Carleman operator.
The image of this Carleman operator is the $\rkhs$ span by $h_j(.)$ since:
$$
\begin{array}{llll}
\forall f \in L^2, \quad (Tf)(x)  &=& \displaystyle \langle \Gamma_x , f \rangle_{L^2} &\\
                             &=& \displaystyle \langle \sum_{j \in J} h_j(x) e_j, \sum_{i \in I} 
\alpha_i e_i\rangle_{L^2} & \mbox{ because } \displaystyle f =\sum_{i \in I} \alpha_i e_i\\
                             &=& \displaystyle \sum_{j \in J}   h_j(x) \sum_{i \in I}  \alpha_i  \langle e_j,e_i\rangle_{L^2} \\
                             &=& \displaystyle \sum_{j \in J}   \alpha_j h_j(x) 
\end{array}
$$
and family $\{h_i(.)\}$ is orthonormal since $h_i(.) = T e_i$.
\end{itemize}
\end{proof}

\noindent
To put this framework at work the relevant function $\Gamma_x$ has to be found.
Some examples with popular kernels illustrating this definition are shown table \ref{table:gamma}.

\begin{table}
\centering

\begin{tabular}{| l | c | c |} \hline
Name  & $\displaystyle \Gamma_x(u)$  &  $\displaystyle K(x,y)$   \\ \hline
 Cameron Martin   &  $\displaystyle  \Un^{ }_{\{x \leq u\}}$ & $\min{(x,y)}$   \\
 Polynomial  & $\displaystyle   e_0(u) + \sum_{i=1}^d x_i e_i(u)  $&  $\x^\top \y+1$ \\
 Gaussian  &  $\displaystyle 1/Z exp^{-\frac{(x-u)^2}{2}}$ &  
$\displaystyle 1/Z' exp^{-\frac{(x-y)^2}{4}}$ \\ \hline
\end{tabular}
\caption{Examples of Carleman operator and their associated reproducing kernel. 
Note that functions $\{e_i\}_{i=1,d}$ are a finite subfamily of a $L^2$ orthonormal basis. $Z$ and $Z'$ are two constants.
} \label{table:gamma}
\end{table}
%

\subsection{Carleman operator and the regularization operator}

The same kind of operator has been introduced by Poggio and Girosi in the regularization framework \cite{poggio89theory}.
They proposed to define the regularization term $\Omega(f)$  (defined equation \ref{eq:cout})
by introducing a regularization operator $P$ from  hypothesis set $\clF$ 
to $L^2$ such that $\Omega(f) = \|Pf\|_{L^2}^2$.
This framework is very attractive since  operator $P$ 
models the prior knowledge about the solution defining its regularity in terms of
derivative or Fourier decomposition properties.
Furthermore the authors show that, in their framework, 
the solution of the learning problem is a linear combination of a kernel (a representer theorem).
They also give a methodology  to build this kernel  as the green function of a differential operator.
Following \cite{Aronszajn50} in its introduction the link between green function and $\rkhs$ 
is straightforward when green function is a positive kernel.
But a problem arises when operator $P$ is chosen as a derivative operator 
and the resulting kernel is not derivable 
(for instance when $P$ is the simple derivation,
 the associated kernel is the non-derivable function $\min(x,y)$).
A way to overcome this technical difficulty is to consider  things the other way round 
by defining the regularization term as the norm of the function in the $\rkhs$ built based on Carleman operator $T$.
In this case we have $\Omega(f) = \|f\|_H = \|T^{-1}g\|_{L^2}^2$.
Thus since $T$ is bijective we can define operator $P$ as:
$P$ = $T^{-1}$.
This is no longer a derivative operator but a generalized derivative operator 
where the derivation is defined as the inverse of the integration ($P$ is defined as $T^{-1}$).

\subsection{Generalization}
It is important to notice that the above framework can be generalized to 
non $L^2$ Hilbert spaces. A way to see this is to use Kolmogorov's dilation theorem \cite{KolDilTheorem}. 
Furthermore, the notion of reproducing kernel itself can be generalized to non-pointwise defined function
by emphasizing the role played by  continuity through positive generalized kernels 
called Schwartz or hilbertian kernels \cite{Schwartz64}.
But this is out of the scope of our work.

\section{Reproducing kernel spaces (RKS)}

\noindent
By focusing on the relevant hypothesis for learning we are going 
to generalize the above framework to non-hilbertian spaces.

\subsection{Evaluation  spaces}
\begin{definition}[ES]\label{ES}
$\,$\\
Let $\clF$ be a real topological vector space (t.v.s.) on an arbitrary set $\clX$, 
$\clF\subset \R^{\clX}$.
$\clF$ is an evaluation  space if and only if:
$$
   \forall x \in \clX, \; \begin{array}{rll}
\delta_{x}: \clF &\longrightarrow& \R\\
f &\longmapsto & \delta_{x}(f) = f(x)
\end{array}
 \mbox{ is continuous }$$
\end{definition}

\noindent
ES are then topological vector spaces in which $\delta_t$ (the evaluation functional at  $t$) is continuous,
{\em i.e.} belongs to the topological dual $\clF^{*}$of $\clF$.
\begin{remark}\label{rclX}
Topological vector space $\R^{\clX}$ with the topology of simple convergence 
is by construction an ETS (evaluation topological space).
\end{remark}

\noindent
In the case of normed vector space, another characterization can be given:

\begin{proposition}[normed ES or BES]\label{RKNS}
$\,$\\
Let $\left(\clF,\|.\|_{\clF}\right)$ be a real normed vector space on an arbitrary set 
$\clX$, $ \clF \subset \R^{\clX}$.
$\clF$ is an evaluation kernel space if and only if the evaluation functional:
$$
   \forall x\in \clX, \; \exists M_{x} \in \R, \; \forall f\in \clF, \, |f(x)|\leq M_{x}\|f\|_{\clF}
$$
if it is complete for the corresponding norme it is a Banach evaluation space (BES).
\end{proposition}

 \begin{remark}
In the case of a Hilbert space, we can identify $\clF^{*}$ and $\clF$ 
and, thanks to the Riesz theorem, the evaluation functional can be seen as a function belonging to
$\clF$: it is called the reproducing kernel.
  \end{remark}
This is an important point: thanks to the Hilbertian structure
 the evaluation functional can be seen as a hypothesis function and therefore
 the solution of the learning problem can be built as a linear combination of this reproducing kernel taken different points.
Representer theorem \cite{kimeldorf_wahba} demonstrates this property when the learning machine 
 minimizes a regularized quadratic error criterion.
We shall now generalize these properties to the case when no hilbertian structure is available.

\subsection{Reproducing kernels}

The key point when using Hilbert space is the dot product.
When no such  bilinear positive functional is available its role can be played by a duality map.
Without  dot product, the hypothesis set $\clF$ is no longer in self duality.
We need another set $\clM$ to  put in duality with $\clF$.
This second set $\clM$ is a set of functions measuring
how the information I have  at  point $x_1$ 
helps me to measure the quality of the hypothesis at  point $x_2$.
These two sets have to be in relation through a specific bilinear form.
This relation is called a duality.
\begin{definition}[Duality between two sets]
Two sets $(\clF,\clM)$ are  in duality if 
there exists a bilinear form $\clL$ on $\clF \times \clM$ that separates $\clF$ and $\clM$ (see \cite{Mary1}
for details on the topological aspect of this definition). 
\end{definition}
Let $\clL$ be such a  bilinear form on $\clF \times \clM$ that separate them. 
Then 
we can define a linear application $\gamma_\clF$ and its reciprocal   $\theta_\clF$ as follows:
$$
\begin{array}{lllllcll}
\gamma_\clF  : & \clM &\longrightarrow & \clF^*     
\qquad \qquad  \qquad \qquad  \qquad \qquad  
 \theta_\clF  : & \mbox{Im}\left(\gamma_\clF\right)&\longrightarrow & \clM       \\
            & f          &\longmapsto     & \gamma_\clF f = \clL(.,f)           
            & g=\clL(.,f)   &\longmapsto     & \theta_\clF g = f \\
\end{array}
$$
where $\clF^*$ (resp. $\clM^*$) denotes the dual set of $\clF$ (resp. $\clM$).

\smallskip
\noindent
Let's take an important example of such a duality.
\begin{proposition}[duality of pointwise defined functions]
Let $\clX$ be any set (not necessarily  compact).   $\dbR^\clX$ and $\dbR^{[\clX]}$ are in duality
\end{proposition}
\begin{proof}
Let's define the bilinear application $\clL$ as follows:
$$
\begin{array}{lcll}
 \clL  : & \dbR^\clX \times \dbR^{[\clX]} &\longrightarrow & \dbR       \\
            & \displaystyle \bigl(f(.),g(.)=\sum_{i\in I} \alpha_i \Un_{x_i}(.) \bigr)  
 &\longmapsto     & \displaystyle  \sum_{i\in I}  \alpha_i  f(x_i) = \sum_{x \in \clX} f(x) g(x)\\
\end{array}
$$
\end{proof}

\noindent
Another example is shown in the two following functional spaces:
 $$
\displaystyle L^1 = \left\{f \; \Bigl| \Bigr. \; \int_\clX \; |f| \; d\mu < \infty \right\}
  \qquad   \mbox{ and } \qquad
\displaystyle L^\infty = \left\{f \; \Bigl| \Bigr. \; \operatornamewithlimits{ess\,sup}_{x \in \clX} |f|  < \infty \right\}
$$
where for instance $\mu$ denotes the Lebesgue measure. 
Theses two spaces are put in duality through the following duality map: 
$$
\begin{array}{lrll}
\clL: &L^1 \times L^\infty &\longrightarrow &\dbR     \\
      & f,g &\longmapsto& \clL(f,g) = \displaystyle \int_\clX f \; g \; d\mu 
\end{array}
$$

\begin{definition}[Evaluation subduality]
Two sets $\clF$ and $\clM$ form an evaluation subduality  iff:
\begin{itemize}
\item[-] they are in duality through their duality map $\gamma_\clF$,
\item[-] they both are subsets of $\dbR^\clX$
\item[-] the continuity of the evaluation functional is preserved through:
$$
\mbox{Span}(\delta_x) = \gamma_{\dbR^\clX}\left(\left(\dbR^{\clX}\right)' \right) \subseteq \gamma_\clF(\clM)
\quad \mbox{ and } \quad
\gamma_{\dbR^\clX}\left(\left(\dbR^{\clX}\right)' \right) \subseteq \theta_\clF(\clF)
$$
\end{itemize}
\end{definition}
The key point is the way of preserving the continuity.
Here the strategy to do so is 
first to consider two sets in duality
and then to build the (weak) topology such that the dual elements are (weakly) continuous.

\begin{center}
\begin{figure}
\begin{tabular}{ccc|ccc}
$\quad$ &
\textsl{ Hilbertian case}  &$\;$ &$\quad$& \textsl{General case}& $\qquad$  \\ \hline
& \xymatrix{
 \left(\dbR^{\clX}\right)' \ar@<-0.5ex>[dr]_{i^{*}}  \ar@{.>}@/^1.7pc/[ddrr]^{\varkappa} & \\
& \clF'\stackrel{Riesz}{=}\clF \ar@<0.5ex>[dr]_{i} \\
& & \dbR^\clX }
 &$\;$ &$\quad$
& \xymatrix{
  \left(\dbR^{\clX}\right)' \ar[d]_{i^{*}} \ar[r]^{j^{*}}  
\ar@{.>}@/^1.7pc/[ddrr]_{\varkappa} & \clM'\ar[dr]^{\theta_{\clM}}\\
 \clF'\ar@{>}[dr]_{\theta_{\clF}}  & & \clF\ar[d]^{i}\\
 & \clM \ar[r]_{j} & \dbR^\clX } &\\
&$\boxed{K(s,t)=\langle K(s,.),K(.,t) \rangle_{\clF}}$ &
&& $\boxed{K(s,t)=\clL_{\clF}\left(\varkappa^{*}(\delta_{s}),\varkappa(\delta_{t})\right)}$ \\
&&
\end{tabular}
\caption{illustration of the subduality map.} \label{fig:subdual}
\end{figure}
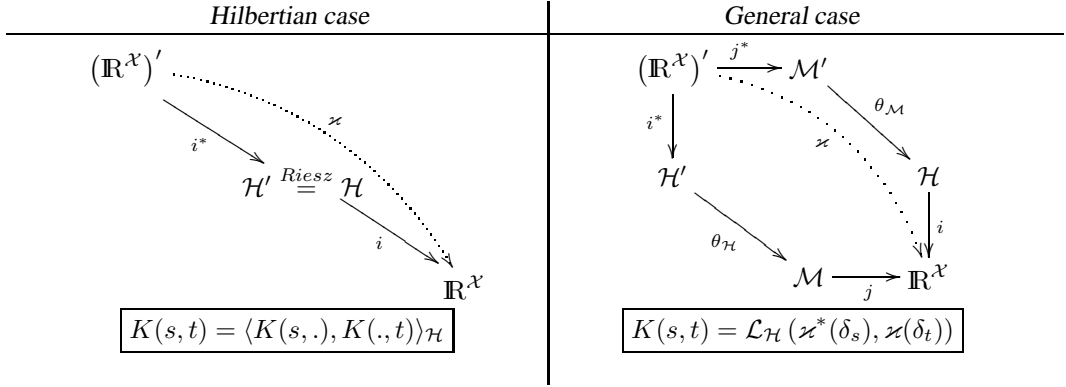
\end{center}

\vspace{-1cm}

\begin{proposition}[Subduality kernel]
A unique weakly continuous linear application $\varkappa$ is associated to each subduality.
This linear application, called the subduality kernel, is defined as follows:
$$
\begin{array}{llll}
\varkappa:  &  \left(\dbR^{\clX}\right)' &\longrightarrow & \dbR^\clX\\
               &  \sum_{i\in I} \delta_{x_i} &\longmapsto     & i \circ \theta_\clM \circ j^* (\sum_{i\in I} \delta_{x_i})\\
\end{array}
$$
where $i$ and $j^*$ are  the canonical injections from $\clF$ to $\dbR^\clX$ and respectively
from $\left(\dbR^{\clX}\right)' $ to $\clM'$ (figure 1).

\end{proposition}
\begin{proof}
for details see \cite{Mary1}.
\end{proof}
We can illustrate this mapping detailing all performed applications as in figure 1:
$$
\begin{array}{ccccccccc}
 \left(\dbR^{\clX}\right)' &\overset{\mbox{see 3.5}}{\longrightarrow} & \dbR^{[\clX]} 
&\overset{j^*}{\longrightarrow} & \clM' 
 &\overset{\theta_\clM }{\longrightarrow} & \clF &\overset{i}{\longrightarrow} & \dbR^\clX\\
 \delta_{x} &\longmapsto     & \Un_{\{x\}}&\longmapsto     & \clL(K_x,.)&\longmapsto     
& K_x(.) &\longmapsto     & K(x,.)\\
\end{array}
$$

\begin{definition}[Reproducing kernel of an evaluation subduality]
Let $(\clF,\clM)$ be an evaluation subduality with respect to  map $\clL_\clF$ 
associated with  subduality kernel $\varkappa$.
The reproducing kernel associated with this evaluation subduality 
is the function of two variables defined as follows:
$$
\begin{array}{llll}
K : & \clX \times \clX &\longrightarrow & \dbR\\
               &  (x,y) &\longmapsto     & 
K(x,y) = \clL_\clF\left(\varkappa^*(\delta_y),\varkappa(\delta_x) \right)_{}\\
\end{array}
$$
\end{definition}
This structure is illustrated in figure \ref{fig:subdual}.
Note that this kernel no longer needs to be definite positive. 
If the kernel is definite positive it is associated with a unique $\rkhs$.
However, as shown in example 8.2.1 it can also be associated with evaluation subdualities.
A way of looking at things is to define $\varkappa$ as the generalization of the Schwartz kernel
while $K$ is the generalization of the Aronszajn kernel to non hilbertian structures.
Based on these definitions the important expression property is preserved.
\begin{proposition}[generation property]

$
\forall f \in \clF, \; \exists (\alpha_i)_{i\in I} \mbox{ such that } \; 
f(x) \approx \sum_{i \in I} \alpha_i K(x,x_i)$ 
and
$\qquad 
\forall g \in \clM, \; \exists (\alpha_i)_{i\in I} \mbox{ such that } \; 
g(x) \approx \sum_{i \in I} \alpha_i K(x_i,x)
$
\end{proposition}
\begin{proof}
This property is due to the density of Span$\{K(.,x), x\in \clX\}$ in $\clF$.
For more details see \cite{Mary1} Lemma 4.3.
\end{proof}

\smallskip
\noindent
Just like $\rkhs$, another important point is the possibility to build an evaluation subduality,
and of course its kernel, starting from any duality.
\begin{proposition}[building evaluation subdualities]
Let $(A,B)$ be a duality with respect to map $\clL_A$.
Let $\{\Gamma_x,x\in \clX\}$ be a total family in $A$ and
 $\{\Lambda_x, x\in \clX\}$ be a total family in $B$.
Let $S$ (reps. $T$) be the linear mapping from $A$ (reps. $B$) to $\dbR^\clX $ 
associated with $\Gamma_x$ (reps. $\Lambda_x$) as follows:
 $$
\begin{array}{llllllll}
S : & A   &\longrightarrow & \dbR^\clX &\qquad\qquad
T : & B   &\longrightarrow & \dbR^\clX \\
    & g   &\longmapsto     & Sg(x)  = \clL_A\left(g,\Lambda_x \right)
 &  & f   &\longmapsto     & Tf(x)  = \clL_A\left(\Gamma_x,f \right)
\end{array}
$$
Then $S$ and $T$ are injective and $(S(A),T(B))$ is an evaluation subduality
with the reproducing kernel $K$ defined by:
$$
K(x,y) = \clL_A(\Gamma_x,\Lambda_y)
$$
\end{proposition}
\begin{proof}
see \cite{Mary1} Lemma 4.5 and proposition  4.6 
\end{proof}

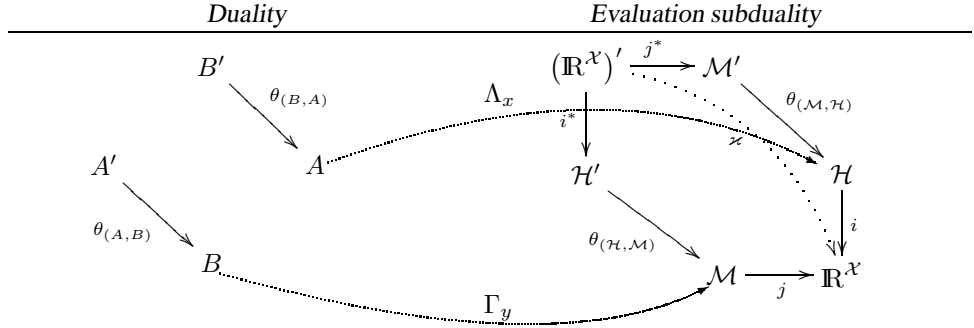
\begin{figure}
\begin{center}
\begin{tabular}{cccccc}
$\quad$ &
\textsl{ Duality }  &$\;$ &$\quad$& \textsl{Evaluation subduality}& $\qquad$  \\ \hline
& \xymatrix{
 & B'\ar[dr]^{\theta_{(B,A)}} \\
A'\ar@{>}[dr]_{\theta_{(A,B)}}& & A \\
& B   & &\\
}
 &$\;$ &$\quad$
& \xymatrix{
  \left(\dbR^{\clX}\right)' \ar[d]_{i^{*}} \ar[r]^{j^{*}}  \ar@{.>}@/^1.7pc/[ddrr]_{\varkappa} & \clM'\ar[dr]^{\theta_{(\clM,\clF)}}\\
 \clF'\ar@{>}[dr]_{\theta_{(\clF,\clM)}}  & & \clF\ar[d]^{i}\\
 & \clM \ar[r]_{j} & \dbR^\clX } &\\
&&
\end{tabular}
\caption{illustration of the building operators for reproducing kernel subduality from a duality $(A,B)$.}
\end{center}

\begin{picture}(165,0)(0,0)

\qbezier[150](110,65)(235,30)(293,59)
\put(291,58){\vector(2,1){4}}

\qbezier[150](151,107)(260,147)(335,107)
\put(331,109){\vector(2,-1){5}}

\put(210,50){\mbox{$\Gamma_y$}}
\put(210,130){\mbox{$\Lambda_x$}}

\end{picture}
\vspace{-1cm}
\end{figure}

\noindent
An example of such subduality is obtained by mapping the $(L^1,L^\infty)$ duality to $\dbR^\clX$ 
using {\em injective} operators defined by the families $\Gamma_x(\tau) = \Un_{\{x <\tau\}}  $ and $\Lambda_y(\tau) = \Un_{\{y <\tau\}}$:
$$
\begin{array}{llll}
T : & L^1 &\longrightarrow & \dbR^\clX \\
    & f   &\longmapsto     & Tf(x)  = \left(\Gamma_x,f \right)_{L^\infty,L^1}=\int \Un_{\{x <\tau\}} f(\tau) \; d\tau
\end{array}
$$
and
$$
\begin{array}{llll}
S : & L^\infty &\longrightarrow & \dbR^\clX \\
    & g   &\longmapsto     & Sg(y)  = \left(g,\Lambda_y \right)_{L^\infty,L^1}=\ \int g(\tau) \Un_{\{y <\tau\}}  \; d\tau
\end{array}
$$
In this case 
$\clF = Im(T)$, $\clM = Im(S)$ 
and
$K(y,x) = \displaystyle \int \Lambda(y,\tau) \Gamma(x,\tau) \;d\tau = \min(x,y)$.
We  define the duality map between $\clF$ and $\clM$ through: $$\displaystyle \clL_\clX(g_1,g_2) = \clL_\clX(Sf_1,Tf_2)  = \clL(f_1,f_2)$$
See example 8.2.1 for details.

\smallskip
\noindent
All useful properties of $\rkhs$ --
pointwise evaluation, continuity of the evaluation functional, representation and building technique --
are preserved.  
A missing dot product has no consequence on this functional aspect of the learning problem.

\section{Representer theorem}

Another issue is of paramount practical importance:  determining the shape of the solution.
To this end representer theorem states that, when $\clF$ is a  $\rkhs$, the solution of the 
minimization of the  regularized cost defined equation (\ref{eq:cout})
%
is a linear combination of the reproducing kernel evaluated at the training  examples 
\cite{kimeldorf_wahba,scholkopf_representer}.
When  hypothesis set $\clF$ is a reproducing space associated with a subduality 
we have the same kind of result.
The solution lies in a finite $n$-dimensional subspace of $\clF$.
But we don't know yet how to systematically build a convenient generating family in this subspace.
\begin{theorem}[representer]
Assume $(\clF,\clM)$ is a subduality of $\; \dbR^\clX$ with kernel $K(x,y)$.
Assume the stabilizer $\Omega$ is convex and differentiable ($\partial_\Omega$ denotes its subdifferential set).

\noindent
If $\partial_\Omega(\sum \alpha_i K(x_i,x)) \subseteq \left\{\sum \beta_i \delta_{x_i}\right\} \in \clH^*$ 
then the solution of cost minimization  lies in 
a $n$-dimensional subspace of $\clF$.
\end{theorem}
\begin{proof}
Define a $\clM$ subset $M_1 = \left\{\sum_{i=1}^n  \alpha_i K(x_i,.)\right\}$.
Let $H_2 \subset \clF$ be the $M_1$ orthogonal in the sense of the duality map 
({\em i.e.} $\forall f\in H_2, \forall g \in M_1 \; \clL(f,g)= 0$). 
Then for all $f \in H_2 ,  f(x_i) = 0, i=1,n$.
Now let $H_1$ be the complement vector space defined such that
$$\clF = H_1 \oplus H_2 \quad \Leftrightarrow \forall f\in \clF \; 
\exists f_1 \in H_1 \mbox{ and } f_2 \in H_2 \quad \mbox{ such that } f = f_1 + f_2
$$ 
The solution of the minimizing problem lies in $H_1$ since:
\begin{itemize}
\item[-] $\forall f_2 \in H_2, C(f_2) = $ constant
\item[-] $\Omega(f_1+f_2) \geq \Omega(f_1) + {\left(\partial_\Omega(f_1) , f_2\right)_{\clM,\clH}} \qquad$
       (thanks to the convexity of $\Omega$)
\item[-] and $\forall f_2 \in H_2, ; {\left(\partial_\Omega(f_1) , f_2\right)_{\clM,\clH}} = 0$ 
       \qquad \quad \;  by hypothesis 
\end{itemize}
By construction $H_1$ a $n$-dimensional subspace of $\clF$.
\end{proof}

\smallskip
\noindent
The nature of  vector space $H_1$ depends on  kernel $K$ and on  regularizer $\Omega$.
In some cases it is possible to be more precise and retrieve the nature of $H_1$. 
Let's assume  regularizer $\Omega(f)$ is given.
$\clF$ may be chosen as the set of function such that $\Omega(f) < \infty$ .
Then, if it is possible to build a subduality $(\clF,\clM)$ with kernel $K$  such that
$$
E = \underbrace{\vect\{K(x_i,.)\}}_{H_1} \oplus  
    \underbrace{\left(\ \vect\{K(.,x_i)\} \right)^\top}_{M_1^\top}
$$ 
and if the vector space spaned by the kernel belongs to the regularizer subdifferential $\partial\Omega(f)$:
$$
\forall f \in H_1, \quad \exists g \in M_1 \mbox{ such that } g \in \partial\Omega(f)
$$
then  solution $f^*$ of the minimization of the regularized empirical cost 
is a linear combination of the kernel:
$$
f^*(x) = \sum_{i=1}^n \alpha_i K(x_i,x)
$$
An example of such result is given with the following regularizer based on the $p$-norm on $G=[0,1]$:
$$
\Omega(f) = \int_0^1 \; \left(f'\right)^p \; d\mu
$$
The hypothesis set is  Sobolev space $H^p $ (the set of functions defined on $[0,1]$ whose generalized derivative is $p$-integrable)
 put in duality with  $H^q $ (with $1/p + 1/q = 1$) 
through the following duality map:
$$
\clL(f,g) = \int_0^1 \; f' g' \; d\mu
$$ 
The associated kernel is just like in  Cameron Martin case $K(x,y) = \min(x,y)$.
Some tedious  derivations lead to:
$$
\forall h \in \clF \quad \clL(h,\partial\Omega(f)) =  \int_0^1 \; h' \; p (f')^{p-1} \; d\mu
$$
Thus the kernel verifies $ p (K(.,y)')^{p-1} \propto K(x,.)$

\smallskip
\noindent
This question of the representer theorem is far from being closed.
We are still looking for a way to derive a generating family from the kernel and the regularizer.
To go more deeply into general and constructive results, a possible way to investigate 
is to go through  $\Omega$ Fenchel dual.



\section{Examples}


\subsection{Examples in Hilbert space}
\noindent
The examples in this section all deal with $\ $\rkhs$ $ included in a $L^{2}$ space. 
\begin{enumerate}
\item
Schmidt ellipsoid:\\
Let $(\clX,\mu)$ be a measure space, $\{e_{i}, i \in I\}$ a basis of $L^{2}(\clX,\mu)$ $I$ being a countable set of indices.
Any sequence $\{\alpha_{i}, i \in I, \quad \sum_{i\in I}\alpha_{i}^{2}<+\infty\}$ 
defines a Hilbert-Schmidt operator on  $L^{2}(\clX,\mu)$ with
kernel function $
    \Gamma(x,y) = \sum_{i \in I}\alpha_{i}e_i(x) e_i(y)
$, thus a reproducing kernel Hilbert space with kernel function:
$$\forall (x,y)\in \clX^{2}, \quad K(x,y)=\sum_{i\in I}\alpha_{i}^{2}e_{i}(x)e_{i}(y)$$
The closed unit ball $\fB_{H}$ of the  $\rkhs$  verifies $$\fB_{H}=T(\fB_{L^{2}})=\left\{f\in L^{2}, 
f=\sum_{i\in I} f_{i}e_{i}, \quad \sum_{i \in I}\left(\frac{f_{i}}{\alpha_{i}}\right)^{2}\leq 1\right\}$$
and is then a Schmidt ellipsoid in $L^{2}$. 
An interesting discussion about  Schmidt ellipsoids and their applications to sample continuity of Gaussian measures may be
found in \cite{Dudley99}.

\item
Cameron-Martin space:\\
Let $T$ be the Carleman integral operator on $L^{2}([0,1]\mu)$ ($\mu$ is the Lebesgue measure) with kernel function
$$
    \Gamma(x,y) = Y(x-y)=\Un_{\{y\leq x\}}
$$
it defines a $\ $\rkhs$ $ 
with reproducing kernel
$ K(x,y)=  \min(x,y)$.
The space $(H;\langle.,.\rangle_{H})$ is the Sobolev space of degree 1, also called the Cameron-Martin space.
$$\left\{
\begin{array}{ll}
H=\{f\; \mbox{absolutely continuous},\exists f'\in L^{2}([0,1]),\; f(x)=\int_{0}^{x}f' d\mu\} \nonumber\\
\langle f,g \rangle_{H}=\langle f',g' \rangle_{L^{2}}                             \nonumber
\end{array}
\right.$$

\item
A Carleman but non Hilbert-Schmidt operator:\\
Let $T$ be the integral operator on $L^{2}(\R,\mu)$ ($\mu$ is the Lebesgue measure) with kernel function
$$
    \Gamma(x,y) = {\exp^{-\frac{1}{2}(x-y)^2}}
$$
It is a Carleman integral operator, thus we can define a $\ $\rkhs$ $ $(H;\langle.,.\rangle_{H})= Im(T)$, but $T$ is not a Hilbert-Schmidt operator.
$H$ reproducing kernel is:
 $$
    K(x,y) = \frac{1}{Z}{\exp^{-\frac{1}{4}(x-y)^2}}
$$
where $Z$ is a suitable constant.
\item
Continuous kernel:\\
This example is based on  theorem 3.11 in \cite{Neveu68}.
Let $\clX$ be a compact subspace of $\R$, $K(.,.)$ a continuous symmetric positive definite kernel. It defines a  $\rkhs$  $(H;\langle.,.\rangle_{H})$ and
any Radon measure $\mu$ of full support is kernel-injective. Then, for any such $\mu$,
there exists a Carleman operator $T$  on $L^{2}(\clX,\mu)$ such that $(H;\langle.,.\rangle_{H})= Im(T)$.

\item
Hilbert space of constants:\\
Let $(H;\langle.,.\rangle_{H})$ be the Hilbert space of constant functions on $\dbR$ 
with scalar product $\langle f,g \rangle_{H}=f(0)g(0)$.
It is obviously a  $\rkhs$  with reproducing kernel $K(.,.)\equiv 1$.
For any probability measure $\mu$ on $\dbR$ let:
$$\forall f\in L^{2}(\R,\mu),\quad T f = \int_{\R}f(s)\mu(ds)
$$
Then $H = T(L^2(\dbR,\mu))$ and $\forall f,g\in H,\;  \langle f,g\rangle_{H}=\langle f,g\rangle_{L^{2}}$.

\item
A non-separable  $\rkhs$  - the $L^{2}$ space of almost surely null functions:\\
Define the positive definite kernel function on $\clX\subset\R$ by $\forall s,t\in \clX,\; K(s,t)=\Un_{\{s=t\}}$. It defines a  $\rkhs$  $(H;\langle.,.\rangle_{H})$
and its functions are null except on a countable set.
Define a measure $\mu$ on  $(\clX,\mB)$ where $\mB$ is the Borel $\sigma$-algebra on $\clX$ by
$\mu(t)=1 \; \forall t\in \clX$. $\mu$ verifies: $\mu(\{t_{1},\cdots,t_{n}\})=n$
and $\mu(A)=+\infty$ for any non-finite $A\in\mB$.  The kernel function is then square integrable and
$H$ is injectively included in
$L^{2}(\clX,\mB,\mu)$. Moreover, $K(s,t)=\int_{\clX}K(t,u)K(u,s)d\mu(u)$ 
with $K$ Carleman integrable and $ T = Id_{L^{2}}$
(note that the identity is a non-compact Carleman integral operator).
Finally, $(H;\langle.,.\rangle_{H})=L^{2}(\clX,\mB,\mu)$.

\item
Separable  $\rkhs$ :\\
Let $H$ be a separable  $\rkhs$ . It is well known that any separable Hilbert space is isomorphic to $\ell^{2}$. 
Then 
there exists
$T$ kernel operator $Im(T)=H$. 
It is easy to construct effectively such a $T$: let $\{h_{n}(.),\, n\in \N\}$ be an orthonormal basis of $H$ 
and define
$T$ kernel operator on $\ell^{2}$ with kernel $\Gamma_x\to \{h_{n}(x),\,n\in \N\}(\in l^{2})$. 
Then $Im(T)=H$.

\end{enumerate}

\subsection{Other examples}

\noindent
Applications to non-hilbertian spaces are also feasible:
\begin{enumerate}

\item
$(L^{1}, L^{\infty})$ - ``Cameron-Martin" evaluation subduality:\\
Let $T$ be the kernel operator on $L^{1}([0,1]\mu)$ ($\mu$ is the Lebesgue measure) with kernel function
$$
    \Gamma(t,s) = Y(t-s)=\Un_{\{s\leq t\}}, \quad  \Gamma(t,.)\in L^{\infty}
$$
it defines an evaluation duality $(H_1;H_\infty)$ with reproducing kernel
$$\forall (s,t)\in \clX^{2}, \quad K(s,t)=  \min(s,t)$$
$$\left\{
\begin{array}{ll}
H_1=\{f\; \mbox{absolutely continuous},\exists f'\in L^{1}([0,1]),\; f(t)=\int_{0}^{t}f'(s)ds\} \nonumber\\
\| f \|_{H_1}=\| f' \|_{L^{1}}                             \nonumber
\end{array}
\right.$$
and 
$$\left\{
\begin{array}{ll}
H_\infty=\{f\; \mbox{absolutely continuous},\exists f'\in L^{\infty}([0,1]),\; f(t)=\int_{0}^{t}f'(s)ds\} \nonumber\\
\| f \|_{H_\infty}=\| f' \|_{L^{\infty}}                             \nonumber
\end{array}
\right.$$


\item
$\left(\R^{\clX}, \R^{[\clX]}\right)$:\\
We have seen that $\R^{\clX}$ endowed with the topology of simple convergence is an ETS.
However, $\R^{\clX}$ endowed with the topology of almost sure convergence is never an ETS unless every singleton of $\clX$ has strictly positive
measure.

\end{enumerate}



%

\section{Conclusion}
It is always possible to learn without kernel.  
But even if it is not visible, one is  hidden somewhere!
We have shown,  from some basic principles 
 (we want to be able to compute the value of a hypothesis at any point
and we want the  evaluation functional to be continuous),
how to derive a framework generalizing  $\rkhs$ to non-hilbertian spaces.
In our reproducing kernel dualities, all  $\rkhs$ nice properties are preserved 
except the dot product replaced by a duality map.
Based on the generalization of the hilbertian case,
it is possible to build associated  kernels thanks to simple operators.
The construction of evaluation subdualities without Hilbert structure
 is easy within this framework (and rather new).
The derivation of evaluation subdualities from any kernel operator has many practical outcome.
First, such operators on separable Hilbert spaces can be represented by matrices, 
and we can build any separable $\rkhs$ from
well-known $\ell^{2}$ structures (like wavelets in a $L^{2}$ space for instance).
Furthermore, the set of kernel operators is a vector space whereas 
the set of evaluation subdualities is not 
(the set of $\rkhs$ is for instance a convex cone), hence
practical combination of such operators are feasible.
On the other hand, from the bayesian point of view, 
this result may have many theoretical and practical implications 
in the theory of Gaussian or Laplacian measures and abstract Wiener spaces.

\smallskip
\noindent
Unfortunately, even if some work has been done, a general representer theorem is not available yet.
We are looking for an automatic  mechanism  designing the {\em shape} 
of the solution of the learning problem in the following way:
$$
\displaystyle f(\x) = \sum_{i=1}^m  \alpha_i K(\x_i,\x) + \displaystyle \sum_{j=1}^k \beta_j \varphi_j(\x)
$$
where  Kernel $K$, number of component $m$  and functions $\varphi_k(\x), j=1,k$ are derivated  from  regularizer $\Omega$. 
The remaining questions being: 
how to learn the coefficients and how to determine  cost function?

\section*{Acknowledgements} Part of this work has been realized while the authors 
were visiting B. Sch\"olkopf in Tuebingen. 
The section dedicated to the representer theorem benefits
from  O. Bousquet ideas. 
This work also benefits from comments and discussion with NATO ASI 
 on  Learning Theory and Practice students in Leuven.


\begin{thebibliography}{10}

{\small
\bibitem{Alpay87}
D.A. Alpay,
\newblock Some krein spaces of analytic functions and an inverse scattering problem,
\newblock {\em Michigan Journal of Mathematics} 
{\bf 34} (1987) 349--359.


\bibitem{Aronszajn50}
N.~Aronszajn.
\newblock Theory of reproducing kernels,
\newblock {\em Transactions of the American Society} 
{\bf 68} (1950) 337--404.


\bibitem{Atteia92}
M. Att{\'e}ia,
\newblock {\em Hilbertian kernels and spline functions},
\newblock North-Holland (1992).


\bibitem{Att81}
M. Att{\'e}ia and J. Audounet,
\newblock Inf-compact potentials and banachic kernels,
\newblock In {\em Banach space theory and its applications}, 
volume 991 of {\em Lecture notes in mathematics}, Springer-Verlag
(1981) 7--27.


\bibitem{CuckerSmale01}
F. Cucker and S. Smale,
\newblock On the mathematical foundations of learning.
\newblock {\em Bulletin of the American Mathematical Society}
{\bf 39} (2002) 1--49.


\bibitem{Dudley99}
R.~M. Dudley,
\newblock {\em Uniform central limit theorems},
\newblock Cambridge university press (1999).


\bibitem{KolDilTheorem}
D.~Evans and J.T. Lewis,
\newblock Dilations of irreversible evolutions in algebraic quantum theory,
\newblock {\em Communications of the Dublin Institute for advanced Studies, Series A.}, 
24 (1977).


\bibitem{girosi_sparse}
F. Girosi,
\newblock An equivalence between sparse approximation and support vector machines,
\newblock {\em Neural Computation} {\bf 10}(6) (1998) 1455--1480.


\bibitem{kimeldorf_wahba}
G.~Kimeldorf and G.~Wahba,
\newblock Some results on {T}chebycheffian spline functions,
\newblock {\em J. Math. Anal. Applic.} {\bf 33} (1971) 82--95.


\bibitem{Mary1}
X. Mary, D. De Brucq and S. Canu,
\newblock Sous-dualit{\'e}s et noyaux (reproduisants) associ{\'e}s,
\newblock Technical report PSI 02-006 (2002).
\newblock available at {\tt asi.insa-rouen.fr/\char126scanu}


\bibitem{Mercer09}
J.~Mercer,
\newblock Functions of positive and negative type and their connection with the
  theory of integral equations,
\newblock {\em Transactions of the London Philosophical Society A}
  {\bf 209} (1909) 415--446.

\bibitem{Neveu68}
J. Neveu,
\newblock {\em Processus al{\'e}atoires gaussiens},
\newblock S{\'e}minaires de math{\'e}matiques sup{\'e}rieures, Les presses de
  l'universit{\'e} de Montr{\'e}al (1968).

\bibitem{poggio89theory}
T. Poggio and F. Girosi,
\newblock A theory of networks for approximation and learning,
\newblock Technical Report AIM-1140 (1989).

\bibitem{Saitoh88}
S. Saitoh,
\newblock {\em Theory of reproducing kernels and its applications}, volume 189.
\newblock Longman scientific and technical (1988).

\bibitem{scholkopf_representer}
B. Sch\"olkopf,
\newblock A generalized representer theorem,
\newblock Technical Report 2000-81, NeuroColt2 Technical Report Series (2000).


\bibitem{Schwartz64}
L. Schwartz,
\newblock Sous espaces hilbertiens d'espaces vectoriels topologiques et noyaux
  associ{\'e}s,
\newblock {\em Journal d'Analyse Math{\'e}matique} (1964) 115--256.


\bibitem{smola98from}
A.J. Smola and B. Sch{\"o}lkopf,
\newblock From regularization operators to support vector kernels,
\newblock In M.I. Jordan, M.J. Kearns, and S.A. Solla, editors,
  {\em Advances in Neural Information Processing Systems}, volume~10. The {MIT}
  Press (1998).

\bibitem{Targon67}
G.I. Targonski,
\newblock On {C}arleman integral operators,
\newblock {\em Proceedings of the American Mathematical Society},
  {\bf 18}(3) (1967) 450--456.

\bibitem{vapnik_nSLT}
V. Vapnik,
\newblock {\em The Nature of Statistical Learning Theory}.
\newblock Springer, N.Y (1995).

\bibitem{wahba_spline}
G. Wahba,
\newblock {\em Spline Models for Observational Data},
\newblock Series in Applied Mathematics, Vol. 59, SIAM, Philadelphia (1990).
}

\end{thebibliography}

\end{document}